\documentclass[journal]{IEEEtran}
\IEEEoverridecommandlockouts
\usepackage{cite}
\usepackage{color,soul}
\usepackage{amsmath,amssymb,amsfonts}
\usepackage{algorithmic}
\usepackage{graphicx}
\usepackage{caption}
\usepackage[colorlinks,
            linkcolor=black,
            anchorcolor=black,
            citecolor=black
            ]{hyperref}
\usepackage{textcomp}
\usepackage{xcolor}
\usepackage{enumerate}
\def\BibTeX{{\rm B\kern-.05em{\sc i\kern-.025em b}\kern-.08em
    T\kern-.1667em\lower.7ex\hbox{E}\kern-.125emX}}

\title{\LARGE \bf
Sensor Fusion for Predictive Control of Human-Prosthesis-Environment Dynamics in Assistive Walking: A Survey
}
\author{Kuangen Zhang$^{1,2}$, Clarence W. de Silva$^{2}$, and Chenglong Fu$^{1,*}$% <-this % stops a space
\thanks{$^{1}$ K. Zhang, C. Fu are with the Department of Mechanical and Energy Engineering, Southern University of Science and Technology, Shenzhen 518055, China (Corresponding author: Chenglong Fu: fucl@sustc.edu.cn).}
\thanks{$^{2}$ K. Zhang, C. W. de Silva are with the Department of Mechanical Engineering, The University of British Columbia, Vancouver V6T1Z4, Canada.}
\thanks{*This work was supported in part by National Natural Science Foundation of China under Grant U1613206, 91648203 and Grant 51335004, and in part by Guangdong Innovative and Entrepreneurial Research Team Program under Grant 2016ZT06G587.}% <-this % stops a space
}

\begin{document}
\captionsetup[figure]{labelformat={default},labelsep=period,name={Figure }}
\captionsetup[table]{labelformat={default},labelsep=period,name={Table }}
\maketitle
\thispagestyle{plain}
\pagestyle{plain}

\begin{abstract}
This survey paper concerns Sensor Fusion for Predictive Control of Human-Prosthesis-Environment Dynamics in Assistive Walking. The powered lower limb prosthesis can imitate the human limb motion and help amputees to recover the walking ability, but it is still a challenge for amputees to walk in complex environments with the powered prosthesis. Previous researchers mainly focused on the interaction between a human and the prosthesis without considering the environmental information, which can provide an environmental context for human-prosthesis interaction. Therefore, in this review, recent sensor fusion methods for the predictive control of human-prosthesis-environment dynamics in assistive walking are critically surveyed. In that backdrop, several pertinent research issues that need further investigation are presented. In particular, general controllers, comparison of sensors, and complete procedures of sensor fusion methods that are applicable in assistive walking are introduced. Also, possible sensor fusion research for human-prosthesis-environment dynamics is presented. 
\end{abstract}

%\begin{IEEEkeywords}
%sensor fusion, Bayesian method, entropy, kalman filter, neural network
%\end{IEEEkeywords}

\section{Introduction}

There were 44, 430 new lower limb amputees in Canada from 2006 to 2011 \cite{imam_incidence_2017}. The situation is more serious in the USA, and it is predicted that the number of persons with limb amputations in the USA will increase to 3.6 million by the year 2050 \cite{ziegler-graham_estimating_2008}.  Without healthy lower limbs, the amputees will face serious difficulties in the daily life. For lower limb amputees, everyday tasks, such as walking, running, and climbing stairs could present major challenges. The lower limb prosthesis can help amputees to walk conveniently \cite{liu_development_2016}, and some commercial passive prostheses are currently available \cite{wang_walk_2015}. However, a passive prosthesis cannot provide the necessary active force during walking, and it will take more energy for amputees to walk with the passive prosthesis when compared to people without physical handicaps \cite{gailey_energy_1994}. In this backdrop it can be concluded that powered prostheses are necessary for amputees to gain the ability to walk properly, and some types of powered lower limb prostheses have been developed in response \cite{au_powered_2008, au_powered_2009, sup_upslope_2011, lawson_control_2013}. Hardware systems of them seem to be reliable, but there are still difficulties for amputees to walk in complex environments with the available powered prostheses. In order to help amputees to walk in complex environments, a prosthesis should have the ability to detect its walking environments, predict the intent of the wearer, and walk predictively. This process is similar to predictive driving \cite{yoshihara_autonomous_2017}, which makes the vehicle follow a desired path without subjecting to uncomfortable jerk or sudden acceleration. Predictive walking can facilitate an active prosthesis to switch locomotion modes and plan a smooth trajectory that avoids tumbling \cite{suzuki_intention-based_2007}. Because the control of the prosthesis is involved in the interactions among the human, prosthesis, and the environment, the foundation of predictive walking is the understanding of human-prosthesis-environment dynamics.

\begin{figure}[htbp]
\centerline{\includegraphics[width=6cm]{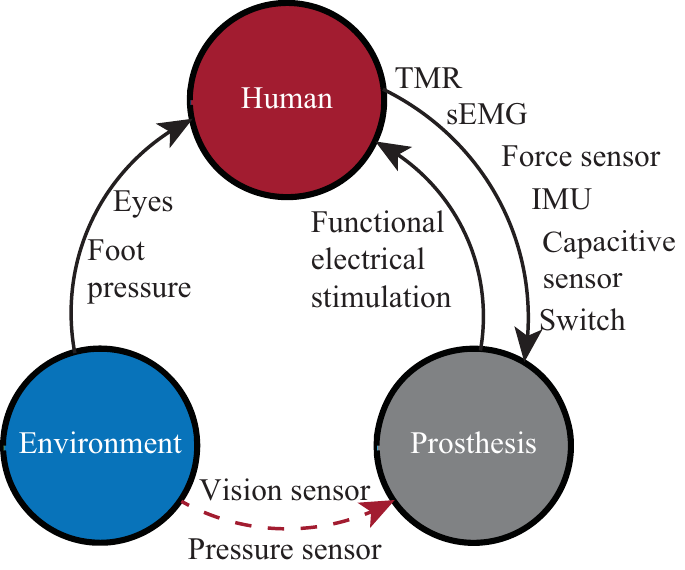}}
\caption{Human-prosthesis-environment dynamics and the corresponding information flow.}
\label{fig:0-HumanProsthesisEnvironmentLoop}
\end{figure}

As shown in \autoref{fig:0-HumanProsthesisEnvironmentLoop} , information flow among the human, prosthesis, and the environment can be realized through different methods. The previous research primarily focused on the problem of the human-prosthesis loop in the context of human intent recognition. Human intent recognition is different from and should precede the problem of human activity recognition. The latter can be performed at high accuracy just using an inertial measurement unit (IMU) \cite{wang_deep_2018}. The human intent happens prior to the motion and is more difficult to be predicted accurately. Although many methods have been introduced to recognize human intent, such as targeted muscle reinnervation (TMR) \cite{souza_advances_2014}, electromyography (EMG) \cite{clites_proprioception_2018}, inertial measurement unit (IMU) \cite{xu_real-time_2018}, and mechanical sensors \cite{simon_configuring_2014}, they are user-dependent and will be affected by the different subjects. Therefore, it is difficult to find a user-independent and reliable method to predict human intent just from the human-prosthesis loop. 

Moreover, vision information can guide people to select optimal paths in different environments \cite{matthis_gaze_2018}, but this vision-locomotion loop is broken in amputees because of the amputation, and it becomes difficult for a prosthesis to understand the human motion intent accurately. Environmental recognition can provide the environmental context of the human motion intent and help the prosthesis to reconstruct the vision-locomotion loop. Consequently, environments should be added to provide the supplementary information, and the human-prosthesis-environment loop will become more complete than the human-prosthesis loop. The predictive control of human-prosthesis-environment dynamics in assistive walking relies on different sensors, so sensor fusion can be used to fuse different information to recognize the walking environment and the human motion intent more accurately and robustly \cite{silva_sensor_2017}.

In order to understand the state of the art better, recent sensor fusion methods for predictive control of human-prosthesis-environment dynamics are presented in this paper. General control methods and corresponding recognition objectives for human-prosthesis-environment dynamics are outlined in \autoref{sec:GeneralControlMethods}. General wearable sensors for realizing the mentioned objectives of assistive walking are indicated in \autoref{sec:GeneralSensors}. Complete procedures of sensor fusion for environmental recognition and human motion intent recognition in the use of an active prosthesis are described in \autoref{sec:FusionMethods}. The conclusion and a general discussion are given in \autoref{sec:Conclusion}.

\section{General control methods}\label{sec:GeneralControlMethods}

\begin{figure*}[htbp]
\centerline{\includegraphics[width=16cm]{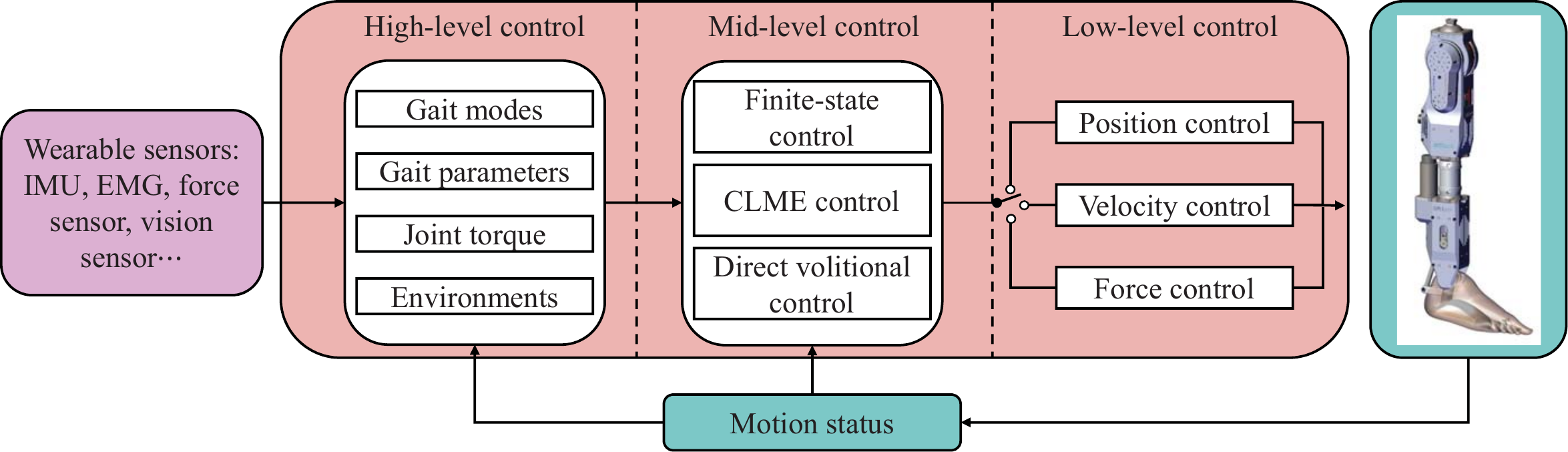}}
\caption{General control architecture of powered lower limb prosthesis.}
\label{fig:1-controlArchitecture}
\end{figure*}

As shown in \autoref{fig:1-controlArchitecture}, the general control architecture for lower limb prosthesis can be divided into high-level, mid-level, and low-level controllers \cite{tucker_control_2015}. This paper focuses on sensor fusion methods for the high-level controller, whose output results are the input of the mid-level controller. Therefore, understanding the popular mid-level controller of lower limb prosthesis can be helpful in determining the objectives of a high-level controller and in selecting suitable sensor fusion methods. Many types of mid-level controllers exist, and we narrow the focus to three most popular mid-level control methods: finite-state control, complementary limb motion estimation (CLME) control, and direct volitional control. These methods are outlined next.

\subsection{Finite-state control}\label{subsec:FSControl}
Finite-state controller is the most popular mid-level controller for powered lower limb prosthesis, which is composed of a series of parametric controllers. Because the human locomotion can be decomposed into different modes
(sitting, standing, walking, up stairs...) \cite{varol_multiclass_2010}, the finite-state controller can use different parameters in different gait modes and environments to control the prosthesis. Moreover, the finite-state controller is usually combined with impedance control \cite{silva_sensors_2015, sup_design_2008}:
\begin{equation}
\tau = k(\theta - \theta_e) + b\Dot{\theta}
\end{equation}
where $\tau$ is the output torque of each joint, and $\theta$ and $\theta_e$ are joint angle and equilibrium joint angle, respectively. Parameters $k$ and $b$ are stiffness coefficient and damping coefficient, respectively.

Because the necessary impedance of the prosthesis should be different in different gait modes and environments, the coefficients stiffness ($k$) and damping ($b$) should be changed appropriately, during control. Existing methods usually tune the parameters manually and save the static parameter values that correspond to specific gait modes and environments. Therefore, this mid-level controller needs a high-level controller to determine the environmental classification and the gait mode prediction of subjects, in order to realize seamless switching between different environments and gait modes.

\subsection{CLME control}\label{subsec:CLMEControl}
Complementary limb motion estimation (CLME) aims to estimate the intended motion of the impaired limb from the residual human body motion \cite{vallery_reference_2009} according to:
\begin{equation}
\begin{bmatrix}
\mathbf{\varphi_p} \\ \mathbf{\Dot{\varphi}_p}
\end{bmatrix}
= \mathbf{K}\begin{bmatrix}
\mathbf{\varphi_h} \\ \mathbf{\Dot{\varphi}_h}
\end{bmatrix} + \mathbf{k}
\end{equation}
Here $\mathbf{\varphi_p}$ and $\mathbf{\Dot{\varphi}_p}$ are the angle vector and the angular velocity vector, respectively, for the prosthesis. The human body motion is described by the joint angle vector $\mathbf{\varphi_h}$ and the joint angular velocity vector $\mathbf{\Dot{\varphi}_h}$, and $\mathbf{K}$ and $\mathbf{k}$ are the mapping matrix and the offset vector, respectively.

This method is similar to echo control, which records the motion of a healthy limb and replays it on an impaired limb \cite{wang_echo-based_2013}. But the CLME can control the prosthesis without time delay and avoid the limitation of symmetric patterns of locomotion. This control method is feasible because of strong inter-joint coordination for human motion \cite{vallery_complementary_2011}. In order to realize this control method, residual body motion should be detected, and the gait parameter prediction and the environmental recognition can be useful as well in predicting the mapping matrix $\mathbf{K}$ and the offset vector $\mathbf{k}$.

\subsection{Direct volitional control}\label{subsec:volitionalControl}
Direct volitional control can directly map residual muscle signals to the motion of prosthesis or change the key parameters of impedance control \cite{ha_volitional_2011}. This method is significant for irregular motion, which is difficult to control using a finite-state controller or the CLME. Although EMG signals are noisy, valid volitional control on the prosthesis has been achieved through EMG signals \cite{huang_voluntary_2018}, and a bidirectional efferent-afferent neural control architecture has been developed \cite{clites_proprioception_2018}. For this mid-level controller, the necessary input is the estimated joint torque of the prosthesis.

Based on the presented discussion, in order to realize reliable mid-level control, the necessary output results of the high-level controller may be the classification of gait modes, recognition of gait parameters, estimation of joint torques, and classification and recognition of the environment.

\section{General sensors}\label{sec:GeneralSensors}
Many sensors such as EMG, IMU, mechanical sensors, and vision sensors are available to realize the mentioned recognition objectives. This paper focuses only on general noninvasive and wearable sensors. Their pros and cons are listed in \autoref{tab:sensors}. More detailed discussion of different wearable sensors can be found in \cite{novak_survey_2015}.

\begin{table*}[htbp]
\centering
\caption {\label{tab:sensors} Signals and comparison of different sensors.}
\renewcommand{\arraystretch}{1.5} % Default value: 1
\begin{tabular}{p{1cm}p{5cm}p{5cm}p{5cm}}
\textbf{Sensor}  &  \textbf{Signals} & \textbf{Pros}  & \textbf{Cons}\\  \hline
IMU
& Angular velocity, acceleration, and direction of magnetic field.
&Stable signal, and the ability to measure joint angles.
&Signals being later than the motion, and being invalid if subjects are still.\\
EMG 
&Electric potential produced by skeletal muscles.
&Signals being prior than the motion, and the ability to predict human intent.
&Low signal-noise ratio, user-dependent signals, and being affected by muscle fatigue.\\
Pressure sensor 
&Ground reactive force signal.
&Reliable signals for detecting gait events.
&Being invalid in the swing phases.\\
Mechanical sensor 
&Joint angle, angular velocity, angular acceleration, and joint torques.
&Reflecting the motion state of the prosthesis.
&Signals being later than human motion, and being difficult to reflect human intent directly.\\
MMG            
&Low-frequency vibrations produced by muscle contraction.
&Being less affected by skin factors. 
&Being affected by muscle fatigue.  \\
SMG
&Changes in muscle structure. 
&Ability to monitor several individual muscles. 
&Being affected by muscle fatigue. \\
EEG
&Motor imagery signals. 
&The ability to detect intended motion in the brain without external stimulus. 
&Large noise, and being difficult to infer the motion intent of lower limb. \\
Vision sensor   
&Point cloud and RGB images of environments.
&Providing environmental context in advance, and being user independent.
&Larger data size, and higher computational complexity.
\\
\end{tabular}
\end{table*}
An IMU consists of a gyroscope, an accelerometer, and a magnetometer, and can be used to detect angular velocity, acceleration, and direction of the earth's magnetic field. These signals are stable and can be fused to estimate the Euler angles of the IMU. However, these signals result from the motion (i.e., the effect of the cause) and cannot be used to predict the intent of the human.

An EMG can measure the electric potential produced by the skeletal muscle. These signals occur prior to the motion, so they are suitable for use in the intent prediction. But the signal-noise ratio of the EMG signal is low and can be affected by the condition of the skin, muscle fatigue, and change of the subject.

A pressure sensor can provide a reliable ground reactive force signal and can be used to detect gait events, such as heel strike and toe off, but it cannot provide valid information in the swing phase.

Mechanical sensors can detect the joint angle, angular velocity, and angular torques of the prosthesis, and can determine the motion status of the prosthesis. However, these signals occur after the motion of the prosthesis and cannot directly reflect the human motion intent.

An MMG is used to detect low-frequency vibration of muscle and is less sensitive to skin conditions when compared with an EMG, but it will be affected by muscle fatigue.

An SMG can measure changes in the muscle structure and can monitor several individual muscles, but it will also be affected by muscle fatigue.

Electroencephalogram (EEG) can be used to detect the intended motion in the brain, and it does not need an external stimulus. But EEG is not accurate and can only be used for classification. Besides, this signal occurs far from the lower limb, and finding an accurate relationship between EEG and the motion of the lower limb may be difficult.

A vision sensor (color camera) may be used to output RGB (red-green-blue) images and the point cloud of environments in front of the prosthesis, and it can provide the environmental context of a human intent. Vision sensor is user-independent, but the data size of the vision sensor is usually large and computational complexity may be higher than in other sensors.

\section{Sensor fusion methods}\label{sec:FusionMethods}
As discussed in \autoref{sec:GeneralControlMethods}, the objectives of the high-level controller of human-prosthesis-environment dynamics can be divided into the classification of the gait modes, recognition of the gait parameters, estimation of the joint torques, and classification and recognition of the environment. These different objectives  can be realized using different sensors and sensor fusion methods. Complete procedure of sensor fusion to realize these objectives are reviewed in this section.

\subsection{Sensor fusion for gait mode classification}
\begin{figure*}[htbp]
\centerline{\includegraphics[width=16cm]{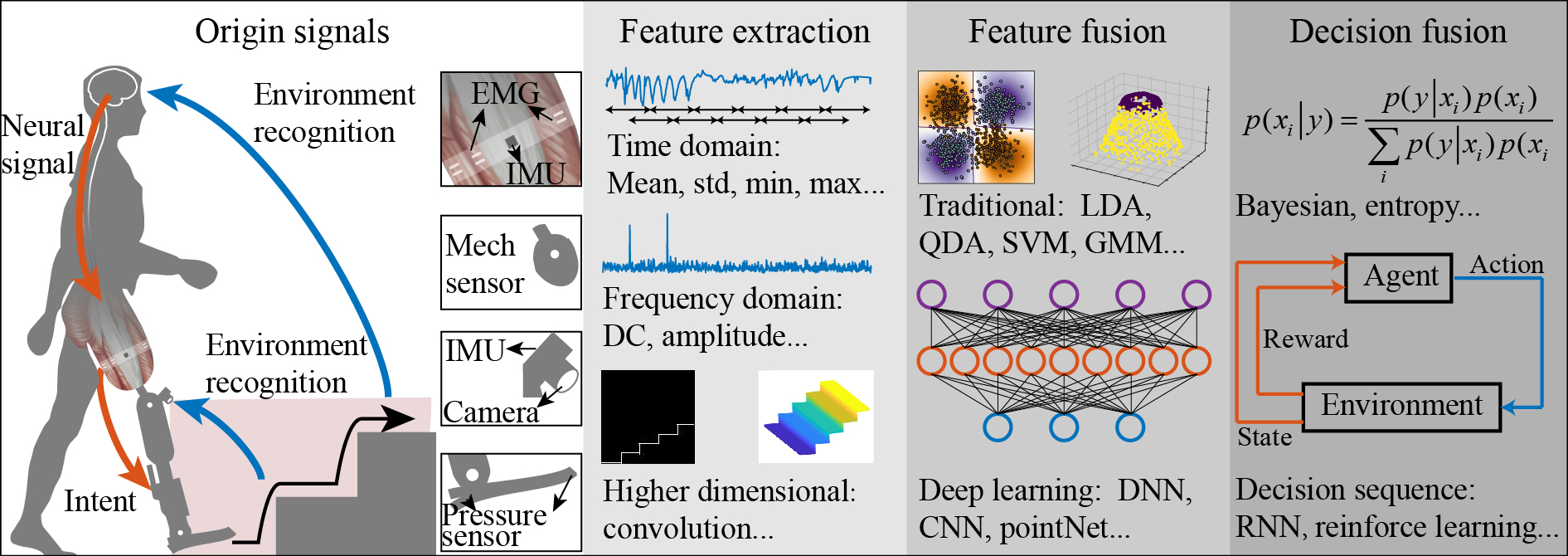}}
\caption{Sensor fusion methods for classification.}
\label{fig:2-sensorfusionForClassification}
\end{figure*}

As stated before, the control parameters of the finite-state controller are different in different gait modes. Therefore, accurate classification of gait modes for the next step is important for predictive control of prosthesis. The complete procedure of gait modes classification is shown in \autoref{fig:2-sensorfusionForClassification}, which can be divided into three parts: feature extraction, feature fusion, and decision fusion.

\subsubsection{Feature extraction}
Signals of most sensors used for gait modes classification are one-dimensional (1D). For 1D signals, the most typical feature extraction method is the extraction of time-domain features. Filtered signals are usually segmented first by sliding windows, and a window may or not overlap with the previous window. Then time-domain features, such as the mean absolute value, the number of zeros crossing, the waveform length, the number of slope sign changes, and auto-regression coefficients are extracted from the precede segmented signals \cite{huang_strategy_2009, huang_continuous_2011, hu_fusion_2018}. Moreover, frequency-domain features may be used as well \cite{hu_fusion_2018, altin_comparison_2016}. A detailed comparison of the features for the classification of elbow gesture is found in \cite{altin_comparison_2016}. Based on this work, auto-regression coefficient and mean value of time-domain features are found to provide superior classification accuracy. Furthermore, the time-domain features do not need signal transformation and can provide fast response. Consequently,  time-domain features are generally better than frequency-domain features. 

Extracted features are the input of the classifiers, and an easily separable feature can simplify the classification algorithm. Thus, some intuitive features have been used to classify gait modes, including the joint translation of leg \cite{stolyarov_translational_2018}, a closed loop of the thigh angle and integration of the thigh angle \cite{bartlett_phase_2018}, and 3D points based on the IMU angle, angular velocity, and acceleration \cite{ledoux_inertial_2018}. Although these features can help in achieving high accuracy of classification, the delayed IMU signal can pose difficulties in the prediction of the gait modes in the next step.

The mentioned features are shallow features and are based on experience, which may be invalid for a different application \cite{wang_deep_2018}. This problem may be solved by using deep learning methods using which deep features can be extracted automatically by the associated convolution layers \cite{dehzangi_imu-based_2017} and max-pooling layers \cite{qin_deep_2018}. Nevertheless, the computation complexity of deep learning methods should be considered because the human intent prediction algorithm should be implemented online on the control board of the prosthesis.

\subsubsection{Feature fusion}\label{subsubsec:featureFusion}
Traditional feature fusion methods that classify gait modes are linear discriminant analysis (LDA) \cite{joshi_terrain_2016}, quadratic discriminant analysis (QDA) \cite{xu_real-time_2018}, support vector machines (SVMs) \cite{zheng_noncontact_2017}, and artificial neural networks (ANNs) \cite{islam_detection_2016}. LDA and QDA are suitable for online implementation because they can calculate analytic optimal parameters, but they are not suitable for features that cannot be separated in lower dimensional spaces. SVM is effective in high dimensional space, but the performance of an SVM relies on the selection of the kernel function. ANN can achieve high accuracy by using a large training set, but it is less computationally efficient than LDA, QDA, and SVM. In a real application, the accuracy and computational complexity should be properly balanced.

Compared to the traditional classification methods, deep learning methods such as a convolutional neural network (CNN) \cite{dehzangi_imu-based_2017, qin_deep_2018}, the voxNet \cite{maturana_voxnet:_2015}, and the pointNet \cite{qi_pointnet++:_2017}, can be more robust and do not rely on the human experience to select suitable kernel functions \cite{wang_deep_2018}. However, the computational complexity of deep learning methods is greater than that of traditional classification methods. Therefore, the number of layers and neurons should not be too high.

\subsubsection{Decision fusion}
Based on the classification results from different sensors and classifiers, the decision fusion can be implemented to get an optimal classification result. Bayesian method \cite{silva_sensor_2017, martinez-hernandez_adaptive_2018} and entropy-based methods \cite{liu_development_2016} can be used to make an optimal decision. The majority voting scheme is usually used to filter the final decision sequences \cite{massalin_user-independent_2018}, and recurrent neural network (RNN) and reinforcement learning methods can also be used for the optimization of the decision sequences.

\subsection{Sensor fusion for environmental classification}
Besides the sensing of human motion, environmental classification is also useful. It has been stated that the human motion is related to human vision \cite{matthis_gaze_2018}. Hence, environmental classification can provide helpful prior information for the prediction of human gait modes.

\subsubsection{Feature extraction}
For the environmental classification, time-domain features (mean, standard deviation, maximum, and minimum of pixels) can be extracted first from sub-regions of depth image or from the local region of point cloud \cite{massalin_user-independent_2018, walas_terrain_2015}. Since this method needs to select a suitable feature, a better approach is to extract deep features based on convolution layers \cite{qin_deep_2018, zhang_environmental_2019}, which can learn features automatically and efficiently.

\subsubsection{Feature fusion}
Environmental classification has been combined with human gait modes classification using the feature fusion methods decision tree \cite{liu_development_2016} and SVM \cite{massalin_user-independent_2018}. Still traditional methods achieve high accuracy for several types of indoor environments such as level ground, up/down stairs, up/down the ramp. However, they may be less robust in complex outdoor environments. Therefore, deep learning methods, including CNN \cite{qin_deep_2018}, 3DCNN \cite{ji_3d_2013}, and pointNet \cite{qi_pointnet++:_2017, zhang_directional_2019}, can be considered for the classification of complex environments, such as pebbled ground and sandy ground. 

\subsection{Sensor fusion for gait parameter recognition.}
Gait parameters (stride length, stride width, foot angle, etc.) recognition can be used to plan the path of the prosthesis in advance, which can be realized based on the previous motion of healthy limb in steady state. The complete procedure of gait parameters recognition can be divided into feature extraction and feature fusion, which are outlined now.

\subsubsection{Feature extraction}
Double integrated acceleration signals of the IMU are usually used for the gait parameter recognition \cite{foxlin_pedestrian_2005}. Since this method may be affected by the zero-velocity phase assumption, it is not suitable for abnormal gait. Therefore, the double pendulum model for the lower limb is preferred for the estimation of stride length \cite{li_wearable_2018}. Recently, convolution layers \cite{hannink_sensor-based_2017} have been used to extract deep features automatically. Pros and cons of these features are similar to those discussed in \ref{subsubsec:featureFusion}.

\subsubsection{Feature fusion}
For feature fusion of gait parameter recognition, state-space models and deep leaning methods have been applied. Different signals can be combined in a state-space model based on geometry and kinematic relationships, and then extend Kalman filter (EKF) \cite{silva_sensor_2017} can be used to estimate gait parameters \cite{foxlin_pedestrian_2005, li_wearable_2018}. Moreover, gait parameters can be also learned by the deep convolution neural network (CNN) directly \cite{ji_3d_2013, hannink_sensor-based_2017}, which does not need a zero velocity phase assumption and can be trained automatically. Typically,  CNN may be computational complex and less efficient than the EKF.

\subsection{Sensor fusion for environmental recognition}
Gait parameter recognition can be based on the previous motion of healthy limb in steady state. However, it is not valid in the transition state (transitions to different environments), and recognized environmental parameters can be used then to predict gait parameters for the next step.

Environmental parameter recognition is usually based on the 3D point cloud. Traditional geometry segmentation algorithm based on the threshold can be used to segment stairs and recognize the size of stairs \cite{krausz_depth_2015}. But deep learning methods, such as pointNet \cite{charles_pointnet:_2017, qi_pointnet++:_2017, qi_frustum_2017}, seem to be more suitable for complex environments. The voxNet needs to integrate volumetric occupancy grid, and it is not suitable for the sparse point cloud \cite{maturana_voxnet:_2015}. The pointNet can realize classification and semantic segmentation of point cloud directly, which is efficient and suitable for online environmental recognition.

\subsection{Sensor fusion for joint torque estimation}
Joint torque estimation is useful in the control of prosthesis for some irregular motions, and acceptable estimation accuracy has been achieved using visual feedback \cite{huang_locomotor_2016}. The complete procedure of joint torque estimation can be divided into feature extraction and feature fusion, which are outlined next.

\subsubsection{Feature extraction}
Joint torque regression usually uses filtered and normalized EMG signals directly \cite{clites_proprioception_2018, huang_voluntary_2018}. A sliding window has been used as well to extract time-domain and frequency-domain features \cite{artemiadis_emg-based_2010}, which is similar to classification.

\subsubsection{Feature fusion}
\begin{figure}[htbp]
\centerline{\includegraphics[width=8cm]{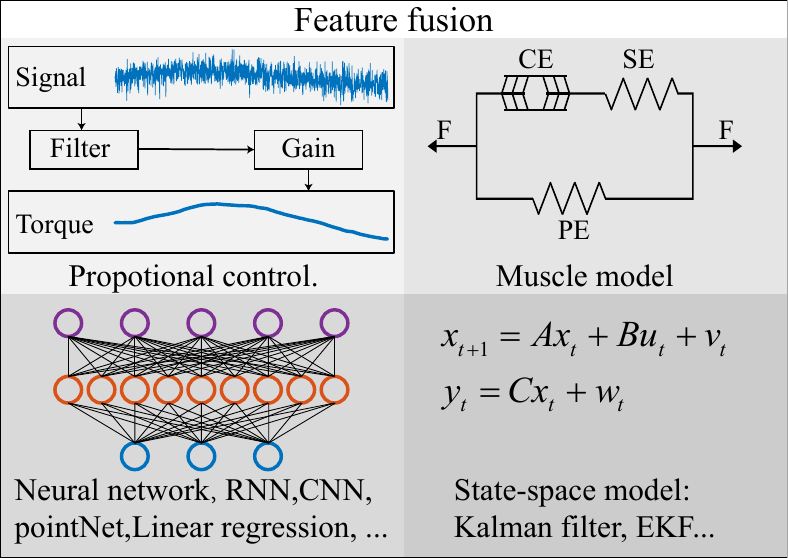}}
\caption{Sensor fusion methods for regression.}
\label{fig:3-sensorfusionForRegression}
\end{figure}

As shown in \autoref{fig:3-sensorfusionForRegression}, feature fusion methods for joint torque regression can be divided into four types: proportional control, muscle model, neural network, and state-space model.

Proportional control is usually used for joint torque estimation from EMG signals \cite{ha_volitional_2011, huang_locomotor_2016, huang_voluntary_2018, clites_proprioception_2018}. For this method, a gain parameter is used to map the filtered and normalized signal to the joint torque, which is convenient to implement. But the quality of the regression result will rely heavily on the quality of the single signal and it is user dependent.

The use of muscle model appears to be a better method, because it considers the variation of joint angle and muscle-tendon dynamics, and should be universal for different subjects \cite{ao_movement_2017, zhuang_admittance_2018}. However, the muscle model method has to calibrate and optimize some parameters for each subject.

Neural networks are also used to estimate joint torques based on the extracted features of EMG \cite{ardestani_human_2014, vujaklija_online_2018}. However, EMG signals are not stationary, and the accuracy will decrease with time if the parameters of the neural network are not updated online. Furthermore, some researchers use state-space models to estimate joint angles \cite{artemiadis_emg-based_2010}. This method is robust to EMG changes caused by muscle fatigue or changes of contraction level. But research on lower limb prosthesis using this method is not readily available, and possible recognition results on amputees are not found.

\section{Conclusion}\label{sec:Conclusion}
This survey paper critically evaluated the past work on Sensor Fusion for Predictive Control of Human-Prosthesis-Environment Dynamics in Assistive Walking, and several pertinent research issues that need further investigation were presented. Previous work has on sensor fusion methods for predictive control of human-prosthesis-environment dynamics in assistive walking were outlined. General sensors, popular mid-level controllers, and complete procedures of sensor fusion methods for predictive control of human-prosthesis-environment dynamics in assistive walking were introduced. Compared to a single sensor, the fusion of different types of sensors can provide more robust and complete information for mid-level controllers. Moreover, suitable stationary features or deep features should be extracted from corresponding sensors, which is beneficial for the generalization of the human motion intent prediction. Additionally, online dataset updating and training are necessary for non-stationary signals. Furthermore, classification and regression are not mutually independent, and their sensor fusion methods can be combined to get better results. 

Further work would be needed to improve the interaction in the human-prosthesis-environment loop. It is proposed that further research should be carried out in the following four areas, in particular.

\subsection{Fusion of environmental information and human intent}
Vision-locomotion loop is very important for human walking in complex environments. Unfortunately, this loop is broken in amputees because of the amputation, and the information flows between an amputee and a prosthesis are user-dependent and not intuitive. Under this circumstance, the ability to make a decision independently for prosthesis should be increased, and the vision-locomotion loop should be reconstructed for a prosthesis to increase environmental adaptability. Moreover, as discussed in section III, typically a single sensor is not adequate to predict human intent. An IMU can recognize human motion but it is available after the motion, which is not useful for controlling that motion. An EMG signal is available prior to the human motion, but it is not stationary and its signal-noise ratio is low. A vision sensor can provide environmental information, but it cannot reflect the human motion intent directly. Therefore, these sensors need to be fused to provide complete information. Further research is needed in this aspect.

\subsection{Stationary deep feature extraction}
Many wearable sensors, such as EMG and EEG, are user dependent and not stationary, which will cause difficulty in the generalization of high-level controllers. Hence, stationary features should be designed to increase the accuracy and robustness of high-level controllers. Lower limb kinematics, such as the inverted pendulum, and muscle-tendon dynamics can be used to extract stationary features from signals of wearable sensors. A deep neural network can be an alternative method to learn deep features automatically, but the network architecture cannot be too complex. Further research is suggested in this area.

\subsection{Online dataset updating and training}
For non-stationary signals, such as EMG, EEG, and MMG, classification and regression accuracy will decrease with time. Therefore, high-level controllers that are trained offline maybe not suitable for long-time use. Because the delayed human activity recognition can be done at high accuracy using IMU or mechanical sensors, leading to history recognition results, which can be used to label the corresponding signals and update the dataset. Then, high-level controllers can be trained online to adapt to signal variation. The resulting high-level controllers should be effective and efficient for training and validation. Further research may be done in this aspect.

\subsection{Fusion of classification and regression results}
Human motion can be decomposed into some primitives such as the flexion and extension of the knee in the sagittal plane. An accurate classification of these motion primitives can be beneficial for the joint torque estimation, which can lead to more robust control than direct proportional control. Further research may be carried out on this aspect as well.

Although this review focused on the operation of a prosthesis, the indicated research is relevant in the field of exoskeleton and human activity recognition. This is because there are many similarities in sensor fusion methods for prosthesis control, exoskeleton control, and human activity recognition.
% Bibliography
\bibliographystyle{IEEEtran}
\bibliography{Sensor Fusion}

% Generated by IEEEtran.bst, version: 1.14 (2015/08/26)
\begin{thebibliography}{10}
\providecommand{\url}[1]{#1}
\csname url@samestyle\endcsname
\providecommand{\newblock}{\relax}
\providecommand{\bibinfo}[2]{#2}
\providecommand{\BIBentrySTDinterwordspacing}{\spaceskip=0pt\relax}
\providecommand{\BIBentryALTinterwordstretchfactor}{4}
\providecommand{\BIBentryALTinterwordspacing}{\spaceskip=\fontdimen2\font plus
\BIBentryALTinterwordstretchfactor\fontdimen3\font minus
  \fontdimen4\font\relax}
\providecommand{\BIBforeignlanguage}[2]{{%
\expandafter\ifx\csname l@#1\endcsname\relax
\typeout{** WARNING: IEEEtran.bst: No hyphenation pattern has been}%
\typeout{** loaded for the language `#1'. Using the pattern for}%
\typeout{** the default language instead.}%
\else
\language=\csname l@#1\endcsname
\fi
#2}}
\providecommand{\BIBdecl}{\relax}
\BIBdecl

\bibitem{imam_incidence_2017}
B.~Imam, W.~C. Miller, H.~C. Finlayson, J.~J. Eng, and T.~Jarus,
  ``\BIBforeignlanguage{en}{Incidence of lower limb amputation in {Canada}},''
  \emph{\BIBforeignlanguage{en}{Can J Public Health}}, vol. 108, no.~4, pp.
  374--380, Nov. 2017.

\bibitem{ziegler-graham_estimating_2008}
K.~Ziegler-Graham, E.~J. MacKenzie, P.~L. Ephraim, T.~G. Travison, and
  R.~Brookmeyer, ``\BIBforeignlanguage{en}{Estimating the {Prevalence} of
  {Limb} {Loss} in the {United} {States}: 2005 to 2050},''
  \emph{\BIBforeignlanguage{en}{Archives of Physical Medicine and
  Rehabilitation}}, vol.~89, no.~3, pp. 422--429, Mar. 2008.

\bibitem{liu_development_2016}
M.~Liu, D.~Wang, and H.~H. Huang, ``Development of an {Environment}-{Aware}
  {Locomotion} {Mode} {Recognition} {System} for {Powered} {Lower} {Limb}
  {Prostheses},'' \emph{IEEE Transactions on Neural Systems and Rehabilitation
  Engineering}, vol.~24, no.~4, pp. 434--443, Apr. 2016.

\bibitem{wang_walk_2015}
Q.~Wang, K.~Yuan, J.~Zhu, and L.~Wang, ``Walk the {Walk}: {A} {Lightweight}
  {Active} {Transtibial} {Prosthesis},'' \emph{IEEE Robotics Automation
  Magazine}, vol.~22, no.~4, pp. 80--89, Dec. 2015.

\bibitem{gailey_energy_1994}
R.~S. Gailey, M.~A. Wenger, M.~Raya, N.~Kirk, K.~Erbs, P.~Spyropoulos, and
  M.~S. Nash, ``Energy expenditure of trans-tibial amputees during ambulation
  at self-selected pace,'' \emph{Prosthetics and orthotics international},
  vol.~18, no.~2, pp. 84--91, 1994.

\bibitem{au_powered_2008}
S.~K. Au and H.~M. Herr, ``Powered ankle-foot prosthesis,'' \emph{IEEE Robotics
  Automation Magazine}, vol.~15, no.~3, pp. 52--59, Sep. 2008.

\bibitem{au_powered_2009}
S.~K. Au, J.~Weber, and H.~Herr, ``Powered {Ankle}–{Foot} {Prosthesis}
  {Improves} {Walking} {Metabolic} {Economy},'' \emph{IEEE Transactions on
  Robotics}, vol.~25, no.~1, pp. 51--66, Feb. 2009.

\bibitem{sup_upslope_2011}
F.~Sup, H.~A. Varol, and M.~Goldfarb, ``Upslope {Walking} {With} a {Powered}
  {Knee} and {Ankle} {Prosthesis}: {Initial} {Results} {With} an {Amputee}
  {Subject},'' \emph{IEEE Transactions on Neural Systems and Rehabilitation
  Engineering}, vol.~19, no.~1, pp. 71--78, Feb. 2011.

\bibitem{lawson_control_2013}
B.~E. Lawson, H.~A. Varol, A.~Huff, E.~Erdemir, and M.~Goldfarb, ``Control of
  {Stair} {Ascent} and {Descent} {With} a {Powered} {Transfemoral}
  {Prosthesis},'' \emph{IEEE Transactions on Neural Systems and Rehabilitation
  Engineering}, vol.~21, no.~3, pp. 466--473, May 2013.

\bibitem{yoshihara_autonomous_2017}
Y.~Yoshihara, Y.~Morales, N.~Akai, E.~Takeuchi, and Y.~Ninomiya, ``Autonomous
  predictive driving for blind intersections,'' in \emph{2017 {IEEE}/{RSJ}
  {International} {Conference} on {Intelligent} {Robots} and {Systems}
  ({IROS})}, Sep. 2017, pp. 3452--3459.

\bibitem{suzuki_intention-based_2007}
K.~Suzuki, G.~Mito, H.~Kawamoto, Y.~Hasegawa, and Y.~Sankai, ``Intention-based
  walking support for paraplegia patients with {Robot} {Suit} {HAL},''
  \emph{Advanced Robotics}, vol.~21, no.~12, pp. 1441--1469, 2007.

\bibitem{wang_deep_2018}
J.~Wang, Y.~Chen, S.~Hao, X.~Peng, and L.~Hu, ``Deep learning for sensor-based
  activity recognition: {A} {Survey},'' \emph{Pattern Recognition Letters},
  Feb. 2018.

\bibitem{souza_advances_2014}
J.~M. Souza, N.~P. Fey, J.~E. Cheesborough, S.~P. Agnew, L.~J. Hargrove, and
  G.~A. Dumanian, ``\BIBforeignlanguage{en}{Advances in {Transfemoral}
  {Amputee} {Rehabilitation}: {Early} {Experience} with {Targeted} {Muscle}
  {Reinnervation}},'' \emph{\BIBforeignlanguage{en}{Current Surgery Reports}},
  vol.~2, no.~5, p.~51, Mar. 2014.

\bibitem{clites_proprioception_2018}
T.~R. Clites, M.~J. Carty, J.~B. Ullauri, M.~E. Carney, L.~M. Mooney, J.-F.
  Duval, S.~S. Srinivasan, and H.~M. Herr,
  ``\BIBforeignlanguage{en}{Proprioception from a neurally controlled
  lower-extremity prosthesis},'' \emph{\BIBforeignlanguage{en}{Science
  Translational Medicine}}, vol.~10, no. 443, p. eaap8373, May 2018.

\bibitem{xu_real-time_2018}
D.~Xu, Y.~Feng, J.~Mai, and Q.~Wang, ``Real-time {On}-board {Recognition} of
  {Continuous} {Locomotion} {Modes} for {Amputees} with {Robotic} {Transtibial}
  {Prostheses},'' \emph{IEEE Transactions on Neural Systems and Rehabilitation
  Engineering}, pp. 1--1, 2018.

\bibitem{simon_configuring_2014}
A.~M. Simon, K.~A. Ingraham, N.~P. Fey, S.~B. Finucane, R.~D. Lipschutz, A.~J.
  Young, and L.~J. Hargrove, ``\BIBforeignlanguage{en}{Configuring a {Powered}
  {Knee} and {Ankle} {Prosthesis} for {Transfemoral} {Amputees} within {Five}
  {Specific} {Ambulation} {Modes}},'' \emph{\BIBforeignlanguage{en}{PLOS ONE}},
  vol.~9, no.~6, p. e99387, Jun. 2014.

\bibitem{matthis_gaze_2018}
J.~S. Matthis, J.~L. Yates, and M.~M. Hayhoe, ``Gaze and the {Control} of
  {Foot} {Placement} {When} {Walking} in {Natural} {Terrain},'' \emph{Current
  Biology}, vol.~28, no.~8, pp. 1224--1233.e5, Apr. 2018.

\bibitem{silva_sensor_2017}
C.~W. de~Silva, \emph{\BIBforeignlanguage{en}{Sensor {Systems}: {Fundamentals}
  and {Applications}}}.\hskip 1em plus 0.5em minus 0.4em\relax Taylor \&
  Francis/CRC Press, Dec. 2017.

\bibitem{tucker_control_2015}
M.~R. Tucker, J.~Olivier, A.~Pagel, H.~Bleuler, M.~Bouri, O.~Lambercy, J.~d.~R.
  Millán, R.~Riener, H.~Vallery, and R.~Gassert, ``Control strategies for
  active lower extremity prosthetics and orthotics: a review,'' \emph{Journal
  of NeuroEngineering and Rehabilitation}, vol.~12, no.~1, p.~1, Jan. 2015.

\bibitem{varol_multiclass_2010}
H.~A. Varol, F.~Sup, and M.~Goldfarb*, ``Multiclass {Real}-{Time} {Intent}
  {Recognition} of a {Powered} {Lower} {Limb} {Prosthesis},'' \emph{IEEE
  Transactions on Biomedical Engineering}, vol.~57, no.~3, pp. 542--551, Mar.
  2010.

\bibitem{silva_sensors_2015}
C.~W. de~Silva, \emph{\BIBforeignlanguage{en}{Sensors and {Actuators}:
  {Engineering} {System} {Instrumentation}, {Second} {Edition}}}.\hskip 1em
  plus 0.5em minus 0.4em\relax CRC Press, Jul. 2015.

\bibitem{sup_design_2008}
F.~Sup, A.~Bohara, and M.~Goldfarb, ``\BIBforeignlanguage{en}{Design and
  {Control} of a {Powered} {Transfemoral} {Prosthesis}},''
  \emph{\BIBforeignlanguage{en}{The International Journal of Robotics
  Research}}, vol.~27, no.~2, pp. 263--273, Feb. 2008.

\bibitem{vallery_reference_2009}
H.~Vallery, E.~H. F.~v. Asseldonk, M.~Buss, and H.~v.~d. Kooij, ``Reference
  {Trajectory} {Generation} for {Rehabilitation} {Robots}: {Complementary}
  {Limb} {Motion} {Estimation},'' \emph{IEEE Transactions on Neural Systems and
  Rehabilitation Engineering}, vol.~17, no.~1, pp. 23--30, Feb. 2009.

\bibitem{wang_echo-based_2013}
W.~J. Wang, J.~Li, W.~D. Li, and L.~N. Sun, ``\BIBforeignlanguage{en}{An
  {Echo}-{Based} {Gait} {Phase} {Determination} {Method} of {Lower} {Limb}
  {Prosthesis}},'' 2013.

\bibitem{vallery_complementary_2011}
H.~Vallery, R.~Burgkart, C.~Hartmann, J.~Mitternacht, R.~Riener, and M.~Buss,
  ``Complementary limb motion estimation for the control of active knee
  prostheses,'' \emph{Biomedizinische Technik/Biomedical Engineering}, vol.~56,
  no.~1, pp. 45--51, 2011.

\bibitem{ha_volitional_2011}
K.~H. Ha, H.~A. Varol, and M.~Goldfarb, ``Volitional {Control} of a
  {Prosthetic} {Knee} {Using} {Surface} {Electromyography},'' \emph{IEEE
  Transactions on Biomedical Engineering}, vol.~58, no.~1, pp. 144--151, Jan.
  2011.

\bibitem{huang_voluntary_2018}
S.~Huang and H.~Huang, ``Voluntary {Control} of {Residual} {Antagonistic}
  {Muscles} in {Transtibial} {Amputees}: {Feedforward} {Ballistic}
  {Contractions} and {Implications} for {Direct} {Neural} {Control} of
  {Powered} {Lower} {Limb} {Prostheses},'' \emph{IEEE Transactions on Neural
  Systems and Rehabilitation Engineering}, vol.~26, no.~4, pp. 894--903, Apr.
  2018.

\bibitem{novak_survey_2015}
D.~Novak and R.~Riener, ``A survey of sensor fusion methods in wearable
  robotics,'' \emph{Robotics and Autonomous Systems}, vol.~73, pp. 155--170,
  Nov. 2015.

\bibitem{huang_strategy_2009}
H.~Huang, T.~A. Kuiken, and R.~D. Lipschutz, ``A {Strategy} for {Identifying}
  {Locomotion} {Modes} {Using} {Surface} {Electromyography},'' \emph{IEEE
  Transactions on Biomedical Engineering}, vol.~56, no.~1, pp. 65--73, Jan.
  2009.

\bibitem{huang_continuous_2011}
H.~Huang, F.~Zhang, L.~J. Hargrove, Z.~Dou, D.~R. Rogers, and K.~B. Englehart,
  ``Continuous {Locomotion}-{Mode} {Identification} for {Prosthetic} {Legs}
  {Based} on {Neuromuscular}–{Mechanical} {Fusion},'' \emph{IEEE Transactions
  on Biomedical Engineering}, vol.~58, no.~10, pp. 2867--2875, Oct. 2011.

\bibitem{hu_fusion_2018}
B.~Hu, E.~Rouse, and L.~Hargrove, ``\BIBforeignlanguage{English}{Fusion of
  {Bilateral} {Lower}-{Limb} {Neuromechanical} {Signals} {Improves}
  {Prediction} of {Locomotor} {Activities}},''
  \emph{\BIBforeignlanguage{English}{Frontiers in Robotics and Ai}}, vol.~5,
  p.~78, Jun. 2018, wOS:000436335200001.

\bibitem{altin_comparison_2016}
C.~Altın and O.~Er, ``\BIBforeignlanguage{en}{Comparison of {Different} {Time}
  and {Frequency} {Domain} {Feature} {Extraction} {Methods} on {Elbow}
  {Gesture}’s {EMG}},'' \emph{\BIBforeignlanguage{en}{European Journal of
  Interdisciplinary Studies}}, vol.~2, no.~3, pp. 35--44, Aug. 2016.

\bibitem{stolyarov_translational_2018}
R.~Stolyarov, G.~Burnett, and H.~Herr, ``Translational {Motion} {Tracking} of
  {Leg} {Joints} for {Enhanced} {Prediction} of {Walking} {Tasks},'' \emph{IEEE
  Transactions on Biomedical Engineering}, vol.~65, no.~4, pp. 763--769, Apr.
  2018.

\bibitem{bartlett_phase_2018}
H.~L. Bartlett and M.~Goldfarb, ``\BIBforeignlanguage{en}{A {Phase} {Variable}
  {Approach} for {IMU}-{Based} {Locomotion} {Activity} {Recognition}},''
  \emph{\BIBforeignlanguage{en}{IEEE Transactions on Biomedical Engineering}},
  vol.~65, no.~6, pp. 1330--1338, Jun. 2018.

\bibitem{ledoux_inertial_2018}
E.~Ledoux, ``Inertial {Sensing} for {Gait} {Event} {Detection} and
  {Transfemoral} {Prosthesis} {Control} {Strategy},'' \emph{IEEE Transactions
  on Biomedical Engineering}, pp. 1--1, 2018.

\bibitem{dehzangi_imu-based_2017}
O.~Dehzangi, M.~Taherisadr, R.~ChangalVala, O.~Dehzangi, M.~Taherisadr, and
  R.~ChangalVala, ``\BIBforeignlanguage{en}{{IMU}-{Based} {Gait} {Recognition}
  {Using} {Convolutional} {Neural} {Networks} and {Multi}-{Sensor} {Fusion}},''
  \emph{\BIBforeignlanguage{en}{Sensors}}, vol.~17, no.~12, p. 2735, Nov. 2017.

\bibitem{qin_deep_2018}
N.~Qin, X.~Hu, and H.~Dai, ``Deep fusion of multi-view and multimodal
  representation of {ALS} point cloud for 3d terrain scene recognition,''
  \emph{ISPRS Journal of Photogrammetry and Remote Sensing}, vol. 143, pp.
  205--212, Sep. 2018.

\bibitem{joshi_terrain_2016}
D.~Joshi and M.~E. Hahn, ``\BIBforeignlanguage{en}{Terrain and {Direction}
  {Classification} of {Locomotion} {Transitions} {Using} {Neuromuscular} and
  {Mechanical} {Input}},'' \emph{\BIBforeignlanguage{en}{Annals of Biomedical
  Engineering}}, vol.~44, no.~4, pp. 1275--1284, Apr. 2016.

\bibitem{zheng_noncontact_2017}
E.~Zheng and Q.~Wang, ``Noncontact {Capacitive} {Sensing}-{Based} {Locomotion}
  {Transition} {Recognition} for {Amputees} {With} {Robotic} {Transtibial}
  {Prostheses},'' \emph{IEEE Transactions on Neural Systems and Rehabilitation
  Engineering}, vol.~25, no.~2, pp. 161--170, Feb. 2017.

\bibitem{islam_detection_2016}
M.~Islam and E.~T. Hsiao-Wecksler, ``\BIBforeignlanguage{en}{Detection of
  {Gait} {Modes} {Using} an {Artificial} {Neural} {Network} during {Walking}
  with a {Powered} {Ankle}-{Foot} {Orthosis}},'' 2016.

\bibitem{maturana_voxnet:_2015}
D.~Maturana and S.~Scherer, ``{VoxNet}: {A} 3d {Convolutional} {Neural}
  {Network} for real-time object recognition,'' in \emph{2015 {IEEE}/{RSJ}
  {International} {Conference} on {Intelligent} {Robots} and {Systems}
  ({IROS})}, Sep. 2015, pp. 922--928.

\bibitem{qi_pointnet++:_2017}
C.~R. Qi, L.~Yi, H.~Su, and L.~J. Guibas, ``{PointNet}++: {Deep} {Hierarchical}
  {Feature} {Learning} on {Point} {Sets} in a {Metric} {Space},'' in
  \emph{Advances in {Neural} {Information} {Processing} {Systems} 30},
  I.~Guyon, U.~V. Luxburg, S.~Bengio, H.~Wallach, R.~Fergus, S.~Vishwanathan,
  and R.~Garnett, Eds.\hskip 1em plus 0.5em minus 0.4em\relax Curran
  Associates, Inc., 2017, pp. 5099--5108.

\bibitem{martinez-hernandez_adaptive_2018}
U.~Martinez-Hernandez and A.~A. Dehghani-Sanij, ``Adaptive {Bayesian} inference
  system for recognition of walking activities and prediction of gait events
  using wearable sensors,'' \emph{Neural Networks}, vol. 102, pp. 107--119,
  2018.

\bibitem{massalin_user-independent_2018}
Y.~Massalin, M.~Abdrakhmanova, and H.~A. Varol, ``User-{Independent} {Intent}
  {Recognition} for {Lower} {Limb} {Prostheses} {Using} {Depth} {Sensing},''
  \emph{IEEE Transactions on Biomedical Engineering}, vol.~65, no.~8, pp.
  1759--1770, Aug. 2018.

\bibitem{walas_terrain_2015}
K.~Walas, ``\BIBforeignlanguage{en}{Terrain {Classification} and {Negotiation}
  with a {Walking} {Robot}},'' \emph{\BIBforeignlanguage{en}{Journal of
  Intelligent \& Robotic Systems}}, vol.~78, no.~3, pp. 401--423, Jun. 2015.

\bibitem{zhang_environmental_2019}
K.~Zhang, C.~Xiong, W.~Zhang, H.~Liu, D.~Lai, Y.~Rong, and C.~Fu,
  ``Environmental features recognition for lower limb prostheses toward
  predictive walking,'' \emph{IEEE transactions on neural systems and
  rehabilitation engineering}, 2019.

\bibitem{ji_3d_2013}
S.~Ji, W.~Xu, M.~Yang, and K.~Yu, ``3d {Convolutional} {Neural} {Networks} for
  {Human} {Action} {Recognition},'' \emph{IEEE Transactions on Pattern Analysis
  and Machine Intelligence}, vol.~35, no.~1, pp. 221--231, Jan. 2013.

\bibitem{zhang_directional_2019}
K.~Zhang, J.~Wang, and C.~Fu, ``Directional {PointNet}: 3d {Environmental}
  {Classification} for {Wearable} {Robotics},'' \emph{arXiv:1903.06846 [cs]},
  Mar. 2019, arXiv: 1903.06846.

\bibitem{foxlin_pedestrian_2005}
E.~Foxlin, ``Pedestrian tracking with shoe-mounted inertial sensors,''
  \emph{IEEE Computer Graphics and Applications}, vol.~25, no.~6, pp. 38--46,
  Nov. 2005.

\bibitem{li_wearable_2018}
G.~Li, T.~Liu, and J.~Yi, ``Wearable {Sensor} {System} for {Detecting} {Gait}
  {Parameters} of {Abnormal} {Gaits}: {A} {Feasibility} {Study},'' \emph{IEEE
  Sensors Journal}, vol.~18, no.~10, pp. 4234--4241, May 2018.

\bibitem{hannink_sensor-based_2017}
J.~Hannink, T.~Kautz, C.~F. Pasluosta, K.-G. Gasmann, J.~Klucken, and B.~M.
  Eskofier, ``\BIBforeignlanguage{en}{Sensor-{Based} {Gait} {Parameter}
  {Extraction} {With} {Deep} {Convolutional} {Neural} {Networks}},''
  \emph{\BIBforeignlanguage{en}{IEEE Journal of Biomedical and Health
  Informatics}}, vol.~21, no.~1, pp. 85--93, Jan. 2017.

\bibitem{krausz_depth_2015}
N.~E. Krausz, T.~Lenzi, and L.~J. Hargrove, ``Depth {Sensing} for {Improved}
  {Control} of {Lower} {Limb} {Prostheses},'' \emph{IEEE Transactions on
  Biomedical Engineering}, vol.~62, no.~11, pp. 2576--2587, Nov. 2015.

\bibitem{charles_pointnet:_2017}
R.~Q. Charles, H.~Su, M.~Kaichun, and L.~J. Guibas, ``{PointNet}: {Deep}
  {Learning} on {Point} {Sets} for 3d {Classification} and {Segmentation},'' in
  \emph{2017 {IEEE} {Conference} on {Computer} {Vision} and {Pattern}
  {Recognition} ({CVPR})}, Jul. 2017, pp. 77--85.

\bibitem{qi_frustum_2017}
C.~R. Qi, W.~Liu, C.~Wu, H.~Su, and L.~J. Guibas, ``Frustum {PointNets} for 3d
  {Object} {Detection} from {RGB}-{D} {Data},'' \emph{arXiv:1711.08488 [cs]},
  Nov. 2017, arXiv: 1711.08488.

\bibitem{huang_locomotor_2016}
S.~Huang, J.~P. Wensman, and D.~P. Ferris, ``Locomotor {Adaptation} by
  {Transtibial} {Amputees} {Walking} {With} an {Experimental} {Powered}
  {Prosthesis} {Under} {Continuous} {Myoelectric} {Control},'' \emph{IEEE
  Transactions on Neural Systems and Rehabilitation Engineering}, vol.~24,
  no.~5, pp. 573--581, May 2016.

\bibitem{artemiadis_emg-based_2010}
P.~K. Artemiadis and K.~J. Kyriakopoulos, ``An {EMG}-{Based} {Robot} {Control}
  {Scheme} {Robust} to {Time}-{Varying} {EMG} {Signal} {Features},'' \emph{IEEE
  Transactions on Information Technology in Biomedicine}, vol.~14, no.~3, pp.
  582--588, May 2010.

\bibitem{ao_movement_2017}
D.~Ao, R.~Song, and J.~Gao, ``Movement {Performance} of {Human}–{Robot}
  {Cooperation} {Control} {Based} on {EMG}-{Driven} {Hill}-{Type} and
  {Proportional} {Models} for an {Ankle} {Power}-{Assist} {Exoskeleton}
  {Robot},'' \emph{IEEE Transactions on Neural Systems and Rehabilitation
  Engineering}, vol.~25, no.~8, pp. 1125--1134, Aug. 2017.

\bibitem{zhuang_admittance_2018}
Y.~Zhuang, S.~Yao, C.~Ma, and R.~Song, ``Admittance {Control} {Based} on
  {EMG}-driven {Musculoskeletal} {Model} {Improves} the {Human}-robot
  {Synchronization},'' \emph{IEEE Transactions on Industrial Informatics}, pp.
  1--1, 2018.

\bibitem{ardestani_human_2014}
M.~M. Ardestani, X.~Zhang, L.~Wang, Q.~Lian, Y.~Liu, J.~He, D.~Li, and Z.~Jin,
  ``Human lower extremity joint moment prediction: {A} wavelet neural network
  approach,'' \emph{Expert Systems with Applications}, vol.~41, no.~9, pp.
  4422--4433, Jul. 2014.

\bibitem{vujaklija_online_2018}
I.~Vujaklija, V.~Shalchyan, E.~N. Kamavuako, N.~Jiang, H.~R. Marateb, and
  D.~Farina, ``Online mapping of {EMG} signals into kinematics by
  autoencoding,'' \emph{Journal of NeuroEngineering and Rehabilitation},
  vol.~15, no.~1, p.~21, Mar. 2018.

\end{thebibliography}
\end{document}